\documentclass[letterpaper, 10 pt, conference]{ieeeconf}  

\usepackage{graphicx}
\usepackage{subfig}
\usepackage{booktabs}
\usepackage{cite}
\usepackage{amsmath}
\usepackage{amssymb}

\IEEEoverridecommandlockouts                              

\usepackage{mathtools}

\title{\LARGE \bf
Average-Reward Maximum Entropy Reinforcement Learning for Underactuated Double Pendulum Tasks
}

\author{
Jean Seong Bjorn Choe$^{1}$,
Bumkyu Choi$^{2}$ and Jong-kook Kim$^{3}$
\thanks{$^{1}$School of Electrical Engineering, Korea University, Seoul, South Korea
    {\tt\small garangg@korea.ac.kr}}
\thanks{$^{2}$Independent Researcher
    {\tt\small qjarb3411@korea.ac.kr}}
\thanks{$^{3}$School of Electrical Engineering, Korea University, Seoul, South Korea
    {\tt\small jongkook@korea.ac.kr}}
}

\begin{document}

\maketitle
\thispagestyle{empty}
\pagestyle{empty}

\begin{abstract}
This report presents a solution for the swing-up and stabilisation tasks of the acrobot and the pendubot, developed for the AI Olympics competition at IROS 2024. Our approach employs the Average-Reward Entropy Advantage Policy Optimization (AR-EAPO), a model-free reinforcement learning (RL) algorithm that combines average-reward RL and maximum entropy RL. Results demonstrate that our controller achieves improved performance and robustness scores compared to established baseline methods in both the acrobot and pendubot scenarios, without the need for a heavily engineered reward function or system model. The current results are applicable exclusively to the simulation stage setup. 

\end{abstract}

\section{Introduction}

The AI Olympics competition at IROS 2024\footnote[1]{https://ai-olympics.dfki-bremen.de/} following the 2023 event \cite{ijcai2024p1043} aims to evaluate and compare various control methods in a challenging environment. 

The competition introduces two different configurations of an underactuated double pendulum system\cite{10375556}: the acrobot (inactive shoulder joint) and the pendubot (inactive elbow joint). The task is to develop a controller that can perform an energy-efficient and steady swing-up and then stabilise the robot in the upright position.

The swing-up-and-stabilise task presents interesting challenges for reinforcement learning (RL) based controllers:
\begin{enumerate}
    \item The agent should be penalised for excessive velocity during swing-up, but these penalties could impair exploration or lead to overly rapid swing-ups.
    \item To achieve optimal upright stabilisation, the agent must continue exploring even after finding a suboptimal stable point.
\end{enumerate}

While the most common approach to these challenges with a model-free RL algorithm is to design a hand-crafted reward function, this process is time-consuming and often fails to generalise across different tasks.

In this report, we present a model-free reinforcement learning approach that minimises the need for reward engineering. The primary idea is to formulate the tasks as continuing Markov Decision Processes (MDPs) and introduce an average-reward optimality criterion. Under this setting, the agent is no longer biased toward earlier rewards, as in the typical discounted setting, enabling it to achieve the long-term optimal solution. Furthermore, to learn a robust policy against environmental perturbations and enhance exploration under the constraints, we utilise the maximum entropy RL (MaxEnt RL) framework.

\section{Backgrounds}
\subsection{Average-Reward Reinforcement Learning}
We formulate the problem as a recurrent finite Markov decision process (MDP), consisting of a finite state set $\mathcal{S}$, a finite action set $\mathcal{A}$, a transition probability $p$, and a reward function $r$. For a given start state $s\in\mathcal{S}$ and a stationary policy $\pi: \mathcal{S} \times \mathcal{A} \mapsto [0, 1]$, We define the expected average reward (or the gain) $\rho^\pi$ as:
\begin{align}
    \rho^\pi(s) \coloneqq \left. \lim_{T\rightarrow\infty}\frac{1}{T}\mathbb{E}\left[\sum^{T-1}_{t=0}R_t\right|S_0=s,A_t\sim\pi\right],
\end{align}
where $S_t$ is the state of the MDP at time $t$ and $A_t$ is the action at time $t$, and $R_t$ is the reward received at time $t$.

Average-reward RL \cite{sutton2018reinforcement, dewanto2020average} seeks to find a policy $\pi^*$ that achieves the optimal gain $\rho^{\pi^*}(s)=\sup_\pi \rho^\pi(s)$. For recurrent MDPs, the optimal gain is independent of the start state, i.e., $\rho^{\pi^*} = \rho^{\pi^*}(s), \forall s\in\mathcal{S}$ \cite{puterman2014markov}.

We define the bias value function $v_b^\pi(s)$, which satisfies the Bellman policy expectation equation \cite{dewanto2020average}:
\begin{align}
    v_b^\pi(s) + \rho^\pi = r(s, a) + \sum_{s'\in\mathcal{S}}p(s'|s,a)v_b^\pi(s').
\end{align}
For brevity, we refer to $v_b^\pi(s)$ simply as the value function and use the notation $v^\pi(s)\coloneqq v_b^\pi(s)$.

This approach focuses on optimising the long-term average of rewards, rather than the cumulative discounted rewards. It is particularly useful for continuing tasks where the horizon is indefinite, as it provides a natural way to evaluate policies in ongoing processes without the need for artificial discounting.

\subsection{Maximum Entropy Reinforcement Learning}
Maximum entropy RL (MaxEnt RL) \cite{ziebart2010modeling, haarnoja2018soft, levine2018reinforcement} integrates an entropy term into the objective function, encouraging the agent to maximise both the expected rewards and the entropy of the trajectories simultaneously. In the average-reward setting, the objective becomes:

{
\scriptsize
\begin{align}
   \rho^\pi_\text{soft}(s) \coloneqq \left.\lim_{T\rightarrow\infty}\frac{1}{T}\mathbb{E}\left[\sum^{T-1}_{t=0}R_t - \tau\log\pi(A_t|S_t) \right| S_0=s,A_t\sim\pi\right],
\end{align}
}

where $\tau$ is the temperature hyperparameter controlling the scale of the entropy reward.

MaxEnt RL helps in exploration by encouraging the agent to maintain a diverse set of behaviours and enhances robustness to environmental changes \cite{eysenbach2021maximum}.

\subsection{Entropy Advantage Policy Optimisation}
Entropy Advantage Policy Optimisation (EAPO) \cite{choe2024maximumentropyonpolicyactorcritic} is a model-free, on-policy maximum entropy RL algorithm. The core idea of EAPO is to estimate the reward objective and the entropy objective separately, and then combine them into a soft advantage estimate. Subsequently, an advantage actor-critic algorithm \cite{mnih2016asynchronous} is exploited.

\section{Proposed Method}
We propose Average-Reward Entropy Advantage Policy Optimisation (AR-EAPO), an extension of EAPO to the average reward setting.

We first define the entropy gain $\rho^\pi_\mathcal{H}$ as:
\begin{align}
    \rho^\pi_\mathcal{H} \coloneqq \left. \lim_{T\rightarrow\infty}\frac{1}{T}\mathbb{E}\left[\sum^{T-1}_{t=0}-\tau\log\pi(A_t|S_t) \right| S_0=s,A_t\sim\pi\right],
\end{align}
which is equivalent to the entropy rate of a deterministic MDP, and guaranteed to be bounded for a deterministic recurrent MDP \cite{savas2019entropy}.

The entropy bias function $v^\pi_\mathcal{H}$ satisfies the following Bellman equation:
\begin{align}
    v_\mathcal{H}^\pi(s) + \rho^\pi_\mathcal{H} = -\tau\log\pi(a|s) + \sum_{s'\in\mathcal{S}}p(s'|s,a)v_\mathcal{H}^\pi(s').
\end{align}

Finally, we define the soft bias advantage estimator $\tilde{A}^\pi$ for state and action pairs $(s,a) \in \mathcal{S}\times\mathcal{A}$ as:
\begin{align}
    \tilde{A}^\pi(s,a)&\coloneqq\mathbb{E}_{s'\sim p}[
    \{(r(s,a)-\tau\log\pi(a|s)) - (\rho^\pi + \rho^\pi_\mathcal{H}) \nonumber \\
    &+(v^\pi(s') + v^\pi_\mathcal{H}(s'))\} -(v^\pi(s) + v^\pi_\mathcal{H}(s))] \\
    &\coloneqq\mathbb{E}_{s'\sim p}[\tilde{r}(s,a) + \tilde{v}^\pi(s') - \tilde{v}^\pi(s)] \\
    &\coloneqq A^\pi(s,a) + A^\pi_\mathcal{H}(s,a),
\end{align}
where we also define the soft differential reward function $\tilde{r}$:
\begin{align}
    \tilde{r}(s,a) \coloneqq (r(s,a)-\tau\log\pi(a|s)) - (\rho^\pi + \rho^\pi_\mathcal{H}),
\end{align}
and the soft bias value function $\tilde{v}^\pi$ for states $s\in\mathcal{S}$:
\begin{align}
    \tilde{v}^\pi(s) \coloneqq v^\pi(s) + v^\pi_\mathcal{H}(s),
\end{align}
and the bias advantage functions for each objective:
\begin{align}
    &A^\pi(s,a) \coloneqq \mathbb{E}_{s'\sim p}[r(s,a) + v^\pi(s') -\rho^\pi - v^\pi(s)], \\
    &A^\pi_\mathcal{H}(s,a) \coloneqq \mathbb{E}_{s'\sim p}[-\tau\log\pi(a|s) + v^\pi_\mathcal{H}(s') -\rho^\pi_\mathcal{H} - v^\pi_\mathcal{H}(s)].
\end{align}

The advantages $A^\pi$ and $A^\pi_\mathcal{H}$ are estimated by applying GAE \cite{schulman2015high} for each objective as in EAPO \cite{choe2024maximumentropyonpolicyactorcritic}.

Then, we approximate the gain $\rho^\pi$ and the entropy gain $\rho^\pi_\mathcal{H}$ incrementally using these advantages from a batch of samples \cite{dewanto2020average}:
\begin{align}
    \hat{\rho}^{k+1} = \hat{\rho}^k + \eta\mathbb{E}_t[A^\pi(s_t,a_t)], \\
    \hat{\rho}^{k+1}_\mathcal{H} = \hat{\rho}^k_\mathcal{H} + \eta\mathbb{E}_t[A^\pi_\mathcal{H}(s_t,a_t)],
\end{align}
where $\eta$ is the step size hyperparameter.

Subsequently, the clipped surrogate objective function of Proximal Policy Optimization (PPO) is constructed using the soft bias advantage estimator \cite{schulman2017proximal}:
\begin{align}
    L(\theta) = \hat{\mathbb{E}}_t[\min(r_t(\theta)\tilde{A}^\pi_t,\text{clip}(r_t(\theta),1-\epsilon,1+\epsilon)\tilde{A}^\pi_t],
\end{align}
where $\epsilon$ is the clip range hyperparameter and the expectation is taken over the samples from time steps $t$, and $\theta$ is the parameters of the policy, and $r_t(\theta)$ is the probability ratio $r_t(\theta)\coloneqq\frac{\pi_\theta(a_t|s_t)}{\pi_{\theta_\text{old}}(a_t|s_t)}$, and $\tilde{A}^\pi_t$ is the advantage estimate at time step $t$, i.e., $\tilde{A}^\pi_t \coloneqq \tilde{A}^\pi(s_t,a_t)$.

\section{Environment Design}
To incorporate reinforcement learning into the double pendulum tasks \cite{wiebe2022realaigym}, we follow the approaches used by Zhang et al. \cite{zhang2023solving}, who implemented the Gymnasium \cite{towers_gymnasium_2023} interface for the problem.

\subsection{Observation and Action Space}
We adopt the observation scaling scheme proposed by Zhang et al. \cite{zhang2023solving}, along with the observation normalisation technique using the mean and the standard deviation of observations \cite{andrychowicz2021matters}.

For the action space, we use a 1-dimensional continuous action space as each task allows the agent to control only one joint. We use a control frequency of $100$ Hz for training policies.

\subsection{Reward Function}
We simplify the reward function of Zhang et al. \cite{zhang2023solving} to use only the quadratic reward term:

\begin{equation}
r(\mathbf{s},\mathbf{a}) = - \alpha \left[ (\mathbf{s} - \mathbf{g})^\top Q (\mathbf{s} - \mathbf{g}) + \mathbf{a}^\top R \mathbf{a} \right],
\end{equation}

where:
\begin{itemize}
    \item \( \mathbf{s} \in \mathbb{R}^n \) is the unscaled state vector representing the current observation of the robot.
    \item \( \mathbf{g} = [\pi, 0, 0, 0]^\top \) is the goal state that the robot aims to reach.
    \item \( Q \in \mathbb{R}^{n \times n} \) is a diagonal matrix that determines the amount of penalty for the distance to the goal state.
    \item \( \mathbf{a} \in \mathbb{R}^m \) is the action vector representing the torques applied to the robot's joints.
    \item \( R \in \mathbb{R} \) is a scalar weight for penalising the amount of the applied torque.
    \item $\alpha\in\mathbb{R}^+$ is a scaling factor.
\end{itemize}

We employ an identical reward function for both the acrobot and the pendubot tasks, with $Q=\text{diag}([50,50,4,2])$, $R=1.0$ and $\alpha=0.001$. We observed that AR-EAPO is less sensitive to the choice of $Q$ and $R$ when the scale of the ratio between the reward and the entropy $\left|\frac{r(s,a)}{\tau\log\pi(a|s)}\right|$ is similar.

\subsection{Truncation and Reset}
The environment is reset to a noisy start state $s_0'=s_0 + \varepsilon$, where $\varepsilon$ is a Gaussian noise vector, after 1000 steps (equivalently $10$s) have elapsed.

To alleviate the correlation between vectorised environments, we introduce a novel random truncation method. As we formulate the tasks as continuing tasks, there is no termination condition. Therefore, all parallel environments are at the same time step, inducing a correlation between the samples that harms the on-policy training of AR-EAPO. To address this, we set a small probability $p_\text{trunc}$ of truncating the current "episode" at each time step. This random truncation decorrelates the environments and increases the sample efficiency of the training.

\section{Results}

We used a ReLU network of hidden layers [256, 256] for the policy and a ReLU network of hidden layers [512, 512] for the value network. We trained the policy multiple times for 30 million frames, periodically evaluating its performance and reporting the highest-scoring policy. This process took approximately 100 minutes on a single system equipped with an AMD Ryzen 9 5900X CPU and NVIDIA RTX 3080 GPU. We observed initial convergence at around 20 million frames. Table \ref{table:hyperparmeters} summarises the hyperparameters used for training policies using AR-EAPO.

\begin{table}[h]
    \centering
    \caption{Hyperparmeters used for training.}
    \begin{tabular}{lclc}
        \textbf{Hyperparmeters} & \textbf{Value} & \textbf{Hyperparameters (cont.)} & \textbf{Value} \\
        \midrule
        temperature $\tau$ & 2.0  & log std. init. & -1 \\
        reward GAE $\lambda$ & 0.8  & adv. minibatch norm. & True \\
        entropy GAE $\lambda_e$ & 0.6  &  max grad. norm. & 10.0 \\
        PPO clip range $\epsilon$ & 0.05 & num. envs. & 64 \\
        gain step size $\eta$ & 0.01 & num. rollout steps & 128 \\
        learning rate & $5\mathrm{e}{-4}$ & num. epochs & 6 \\
        $p_\text{trunc}$ & $1\mathrm{e}{-3}$ & batch size & 1024 \\
        EAPO $c_2$& 0.5 & vf. coef. & 0.25 \\
        \bottomrule
    \end{tabular}
    \label{table:hyperparmeters}
\end{table}

\begin{table}[h]
    \centering
    \caption{Performance scores of the Acrobot.}
    \resizebox{0.4866\textwidth}{!}{
    \begin{tabular}{lcccc}
        \textbf{Criteria} & \textbf{AR-EAPO (Ours)} & \textbf{TVLQR} & \textbf{ILQR Riccati} & \textbf{ILQR MPC} \\
        \midrule
        Swingup Success & success & success & success & success\\
        Swingup Time [s] & 1.39 & 4.05 & 4.04 & 4.86 \\
        Energy [J] & 8.32 & 10.43 & 10.55 & 11.54 \\
        Torque Cost [N\(^2\)m\(^2\)] & 1.52 & 1.87 & 1.98 & 2.68 \\
        Torque Smoothness [Nm] & 0.008 & 0.016 & 0.067 & 0.096 \\
        Velocity Cost [m\(^2\)/s\(^2\)] & 117.96 & 105.83 & 106.49 & 110.4 \\
        \textbf{RealAI Score} & \textbf{0.633} & 0.504 & 0.396 & 0.345 \\
        \bottomrule
    \end{tabular}
    }
    \label{table:performance_scores_acrobot}
\end{table}

\begin{table}[h]
    \centering
    \caption{Performance scores of the Pendubot.}
    \resizebox{0.4866\textwidth}{!}{
    \begin{tabular}{lcccc}
        \textbf{Criteria} & \textbf{AR-EAPO (Ours)} & \textbf{TVLQR} & \textbf{ILQR Riccati} & \textbf{ILQR MPC} \\
        \midrule
        Swingup Success & success & success & success & success\\
        Swingup Time [s] & 1.15 & 4.13 & 4.13 & 4.12 \\
        Energy [J] & 7.72 & 9.53 & 9.53 & 9.91 \\
        Torque Cost [N\(^2\)m\(^2\)] & 2.43 & 1.26 & 1.25 & 1.77 \\
        Torque Smoothness [Nm] & 0.0103 & 0.007 & 0.005 & 0.0083 \\
        Velocity Cost [m\(^2\)/s\(^2\)] & 64.326 & 211.12 & 211.34 & 211.98 \\
        \textbf{RealAI Score} & \textbf{0.659} & 0.526 & 0.536 & 0.353 \\
        \bottomrule
    \end{tabular}
    }
    \label{table:performance_scores_pendubot}
\end{table}

Our AR-EAPO controller exhibits significant enhancements compared to the leading baseline controllers for both tasks. In the case of the acrobot, our approach outperforms the TVLQR controller, which stands as the strongest baseline for this task. Similarly, for the pendubot, AR-EAPO surpasses the ILQR Riccati gains controller, the best-performing baseline in this case. The performance improvements are evident across multiple metrics, as detailed in Tables \ref{table:performance_scores_acrobot} and \ref{table:performance_scores_pendubot}.

   \begin{figure}[h]
      \centering
      \includegraphics[width=0.485\textwidth, height=0.45\linewidth]{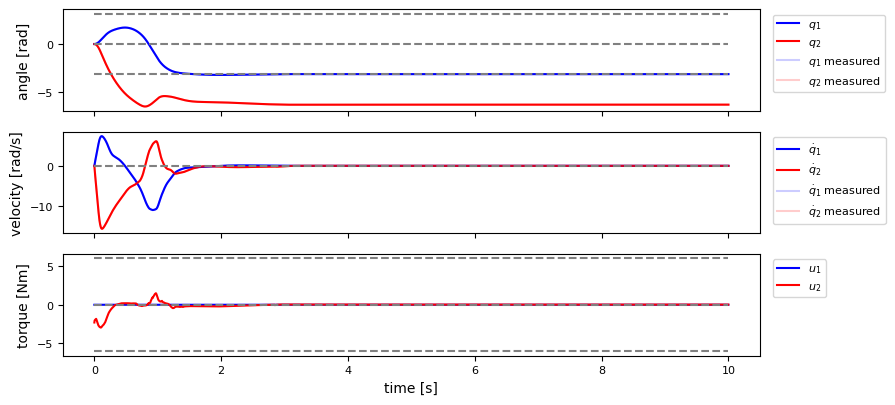}
      \caption{Swing-up trajectory of the acrobot without noise.}
      \label{swing-up_acrobot}
   \end{figure}

  \begin{figure}[h]
      \centering
      \includegraphics[width=0.485\textwidth, height=0.45\linewidth]{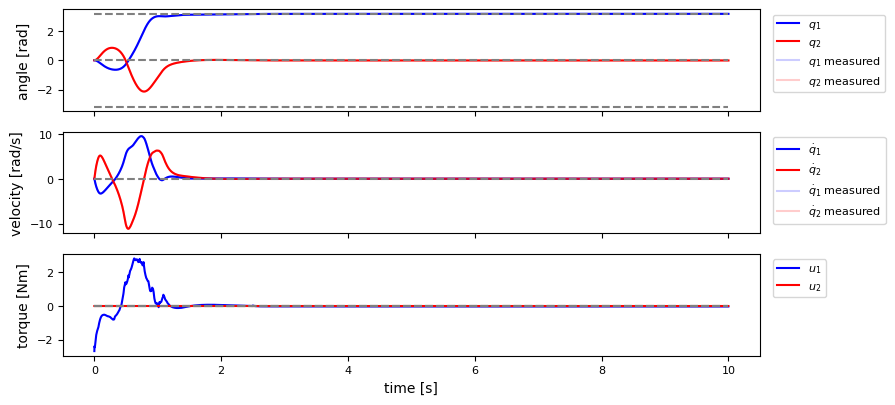}
      \caption{Swing-up trajectory of the pendubot without noise.}
      \label{swing-up_penubot}
   \end{figure}

The time series plots for position, velocity, and torque are shown in Fig. \ref{swing-up_acrobot} for the acrobot and Fig. \ref{swing-up_penubot} for the pendubot.

Regarding robustness, AR-EAPO also yields impressive outcomes. For the acrobot, the overall robustness score is 0.739, surpassing the best baseline (TVLQR) at 0.607. Likewise, for the pendubot, AR-EAPO achieves a score of 0.916 considerably higher than the best baseline (also TVLQR) at 0.767. These scores indicate improved robustness to various perturbations and environmental changes, as further broken down in Tables \ref{table:robust_scores_acrobot} and \ref{table:robust_scores_pendubot}.

\begin{figure}[h]
      \centering
      \includegraphics[width=0.485\textwidth, height=0.45\linewidth]{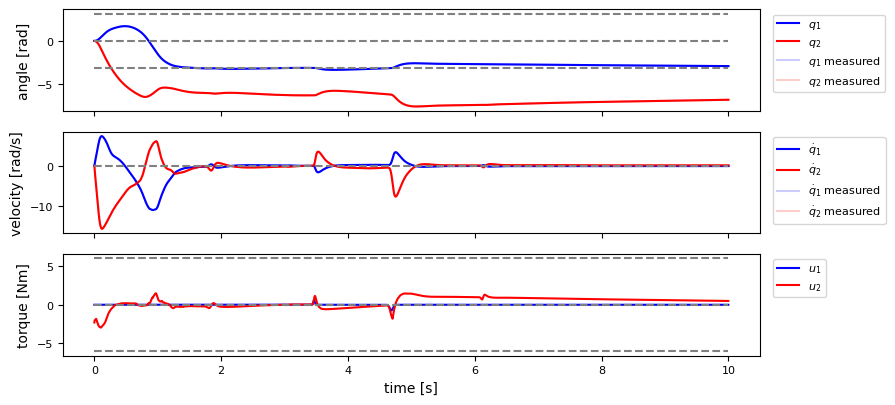}
      \caption{Swing-up trajectory of the acrobot with noise.}
      \label{swing-up_acrobot_robustness}
   \end{figure}

\begin{figure}[h]
      \centering
      \includegraphics[width=0.485\textwidth, height=0.45\linewidth]{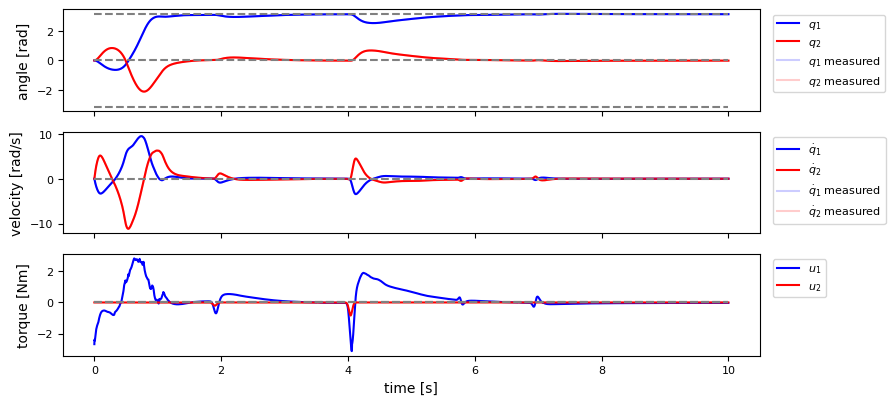}
      \caption{Swing-up trajectory of the pendubot with noise.}
      \label{swing-up_penubot_robustness}
   \end{figure}

\begin{figure}[h]
    \centering
    \subfloat[Acrobot]{\includegraphics[width=0.425\linewidth, height=0.375\linewidth]{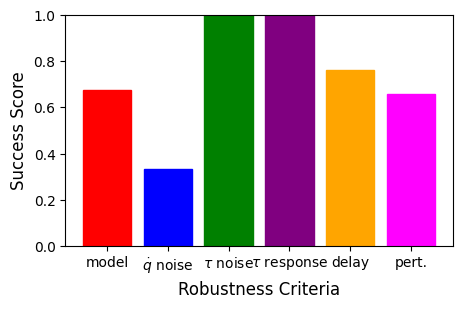}}
    \quad
    \subfloat[Pendubot]{\includegraphics[width=0.425\linewidth, height=0.375\linewidth]{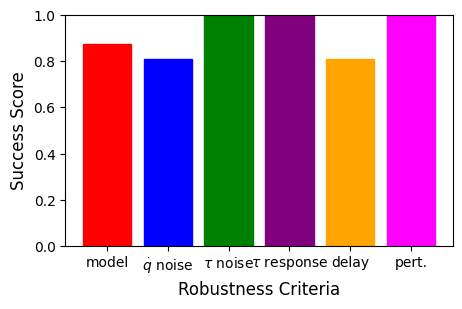}}
    \caption{Robustness results of our controllers for the acrobot and the pendubot.}
    \label{robustness}
\end{figure}

The noisy version of time series plots for position, velocity, and torque are shown in Fig. \ref{swing-up_acrobot_robustness} for the acrobot and Fig. \ref{swing-up_penubot_robustness} for the pendubot.
Fig. \ref{robustness} illustrates these metrics for our controllers in a chart.

\begin{table}[h]
    \centering
    \caption{Robustness scores of the acrobot.}
    \resizebox{0.4866\textwidth}{!}{
    \begin{tabular}{lcccc}
        \textbf{Criteria} & \textbf{AR-EAPO (Ours)} & \textbf{TVLQR} & \textbf{ILQR Riccati} & \textbf{ILQR MPC} \\
        \midrule
        Model [\%] & 67.6 & 48.1 & 4.8 & 6.7 \\
        Velocity Noise [\%] & 33.3 & 19.0 & 9.5 & 9.5 \\
        Torque Noise [\%] & 100.0 & 100.0 & 9.5 & 90.5 \\
        Torque Step Response [\%] & 100.0 & 95.2 & 52.4 & 52.4 \\
        Time delay [\%] & 76.2 & 23.8 & 4.8 & 4.8 \\
        Perturbations [\%] & 66.0 & 78.0 & 2.0 & 42.0 \\ 
        \textbf{Overall Robustness Score} & \textbf{0.739} & 0.607 & 0.138 & 0.343 \\
        \bottomrule
    \end{tabular}
    }
    \label{table:robust_scores_acrobot}
\end{table}

\begin{table}[h]
    \centering
    \caption{Robustness scores of the pendubot.}
    \resizebox{0.4866\textwidth}{!}{
    \begin{tabular}{lcccc}
        \textbf{Criteria} & \textbf{AR-EAPO (Ours)} & \textbf{TVLQR} & \textbf{ILQR Riccati} & \textbf{ILQR MPC} \\
        \midrule
        Model [\%] & 87.6 & 62.9 & 5.2 & 31.9 \\
        Velocity Noise [\%] & 81.0 & 90.5 & 38.1 & 90.5 \\
        Torque Noise [\%] & 100.0 & 100.0 & 9.5 & 100.0 \\
        Torque Step Response [\%] & 100.0 & 100.0 & 85.7 & 100.0 \\
        Time delay [\%] & 81.0 & 42.9 & 14.3 & 38.1 \\
        Perturbations [\%] & 100.0 & 64.0 & 0.0 & 44.0 \\ 
        \textbf{Overall Robustness Score} & \textbf{0.916} & 0.767 & 0.255 & 0.674 \\
        \bottomrule
    \end{tabular}
    }
    \label{table:robust_scores_pendubot}
\end{table}

\section{conclusion}
In this report, we presented a novel solution for the swing-up and stabilisation tasks for the acrobot and the pendubot of the AI Olympics competition at IROS 2024. Our methodology employs the Average-Reward Entropy Advantage Policy Optimization (AR-EAPO), a model-free reinforcement learning algorithm to effectively train a policy capable of both swinging up and stabilising the double pendulum in its upright position.

The results from our experiments indicate that the AR-EAPO controller achieves impressive performance and robustness in both the acrobot and the pendubot scenarios, surpassing the baseline controllers based on optimal control. These promising outcomes highlight the potential of AR-EAPO in handling complex control tasks within the simulated environments. The next step will involve testing and refining our approach in real-robot settings to validate and enhance the applicability of our solution. This transition from simulation to physical implementation will be crucial in fully realising the potential of our method and addressing any challenges that may arise in real-world scenarios.







\bibliographystyle{IEEEtran}
\bibliography{ref.bib}

\end{document}